%% file: acl_latex.tex
\documentclass[11pt]{article}

\usepackage[preprint]{acl}
\usepackage{placeins}
\usepackage{times}
\usepackage{latexsym}

\usepackage[T1]{fontenc}
\usepackage[utf8]{inputenc}

\usepackage{microtype}

\usepackage{inconsolata}

\usepackage{graphicx}

\usepackage{algorithm}
\usepackage{algpseudocode}
\usepackage{amsmath}
\usepackage{booktabs}
\usepackage{enumitem}
\usepackage{colortbl}
\usepackage{arydshln}
\usepackage{multirow}
\usepackage{fvextra}

\DefineVerbatimEnvironment{PromptBox}{Verbatim}{
  fontsize=\scriptsize,
  breaklines=true,
  breakanywhere=true,
  frame=single,
  framesep=2mm,
  baselinestretch=0.95
}

\input{math_commands}

\title{\textsc{MemRefine}: LLM-Guided Compression for Long-Term Agent Memory}

\author{
    Minjae Kim$^{1}$ \; 
    Jinheon Baek$^{2}$ \;
    Soyeong Jeong$^{2}$ \;
    Sung Ju Hwang$^{2,3}$ \\
    $^{1}$Korea University \;\; $^{2}$KAIST \;\; $^{3}$DeepAuto.ai \\
    \texttt{kmj200392@korea.ac.kr} \;
    \texttt{\{jinheon.baek, starsuzi, sungju.hwang\}@kaist.ac.kr}
}

\begin{document}
\maketitle

\input{Sections/1_abstract}
\input{Sections/2_introduction}
\input{Sections/3_related_work}
\input{Sections/4_method}
\input{Sections/5_experimental_setup}
\input{Sections/6_experimental_results}
\input{Sections/7_conclusion}
\input{Sections/8_Limitation}
\input{Sections/9_ethical_considerations}

\bibliography{anthology, custom}

\input{Sections/10_appendix}

\end{document}

%% file: math_commands.tex
\usepackage{amsmath,amsfonts,bm}

\def\eqref#1{equation~\ref{#1}}
\def\1{\bm{1}}

\def\eps{{\epsilon}}

\DeclareMathAlphabet{\mathsfit}{\encodingdefault}{\sfdefault}{m}{sl}
\SetMathAlphabet{\mathsfit}{bold}{\encodingdefault}{\sfdefault}{bx}{n}

\def\gF{{\mathcal{F}}}

\def\gQ{{\mathcal{Q}}}

%% file: Sections/1_abstract.tex
\begin{abstract}
Large language model (LLM) agents are increasingly expected to operate over long-term interactions, where information from past dialogues must be preserved and recalled to support future tasks.
However, as interactions accumulate, the memory store grows without bound and fills with redundant entries that inflate storage cost and degrade retrieval by crowding out the most useful evidence.
Furthermore, this is especially limiting on resource-constrained platforms with hard memory budgets, motivating us to formulate \emph{storage-budgeted memory management}, the task of keeping an already constructed memory store within a fixed budget while preserving information useful for future interactions.
To this end, we then propose \textsc{MemRefine}, an LLM-guided framework that, since surface similarity poorly reflects factual value, uses similarity only to propose candidate pairs and defers delete, merge, and preserve decisions to an LLM judge based on factual content, iterating until the budget is met.
Across multiple memory frameworks and long-term conversation benchmarks, \textsc{MemRefine} consistently meets target budgets while preserving downstream performance and outperforming rule-based baselines under tight budgets.
\end{abstract}

%% file: Sections/2_introduction.tex
\section{Introduction}

\input{Figures/concept}
Large language model (LLM) agents are increasingly expected to operate over long-term interactions, where useful information may be distributed across extended sequences of past conversations, sessions, or events \citep{zhong2024memorybank, maharana2024locomo, xu2025amem}.
To support such interactions, recent agent frameworks equip LLMs with persistent memory stores that record interaction histories and supply them as evidence for future tasks \citep{packer2023memgpt, zhong2024memorybank, xu2025amem, mem0, wu2025longmemeval}.

However, persistent memory creates a storage-efficiency problem that compounds over time.
As the agent continues to interact, the memory store grows without bound and gradually fills with entries that overlap with or duplicate information already captured elsewhere \citep{zhong2024memorybank, xu2025amem, fang2025lightmem, liu2026simplemem}, which both inflates storage cost (a particularly acute issue on resource-constrained platforms such as on-device deployments, where memory budgets are hard-bounded) and degrades retrieval, since the agent may surface near-duplicate or low-value entries.
Motivated by this, we introduce a new task of \emph{storage-budgeted memory management}, where agent memory needs to be continuously kept within a fixed budget while preserving information that remains useful for future interactions.

Existing approaches to compression, however, do not target this setting.
Session-level summarization \citep{wang2023recursive, chen2025comedy} compresses raw conversation history before it enters memory rather than the resulting store, prompt or retrieved-context compression \citep{jiang2023llmlingua, xu2024recomp, jiang2024longllmlingua} operates at inference time on the evidence provided to the LLM, and structure-based graph pruning \citep{brin1998anatomy} relies on criteria such as centrality that do not capture whether entries are factually redundant, complementary, or distinct.
As a result, none of these approaches jointly reduces an already constructed memory store to a target storage budget while reasoning about the factual content of entries.

To address this, we introduce \emph{post-construction memory compression}, where an already constructed memory store is reduced to a target storage budget while retaining information useful for downstream tasks, and propose \textsc{MemRefine}, an LLM-guided framework that instantiates this task as a general module inserted after memory construction and before retrieval, without modifying the host memory pipeline.
The central insight is that while surface-level similarity is useful for surfacing candidate pairs of related memories, similarity alone is too weak to decide their fate, since two textually similar entries may carry redundant facts that can be removed, complementary facts that should be merged, or distinct facts that both warrant preservation.
Motivated by this, \textsc{MemRefine} uses similarity to propose candidate pairs but defers the compression decision itself to an LLM judge, which examines each pair and decides, based on factual content rather than surface wording, whether it should be deleted, merged, or preserved, until the memory store fits within the target storage budget.

We validate \textsc{MemRefine} on two representative memory frameworks with different representations, namely an A-MEM-style graph memory of linked structured notes \citep{xu2025amem} and the Mem0 pipeline that constructs entries through an ingestion-and-update process \citep{mem0}, evaluated across multiple LLM judges and conversation lengths on standard and scaled LoCoMo-style benchmarks \citep{maharana2024locomo}.
Across settings, \textsc{MemRefine} consistently meets the target storage budget while preserving downstream performance under moderate compression, and degrades gracefully rather than collapsing under tighter budgets.
Moreover, compared with rule-based baselines that replace the LLM judge with fixed similarity or graph-structure heuristics, \textsc{MemRefine} is consistently more robust, supporting our hypothesis that compression decisions benefit from LLM-guided factual judgment even when similarity alone is sufficient for candidate selection.

%% file: Figures/concept.tex
\begin{figure}[t]
\centering
\includegraphics[width=0.95\columnwidth]{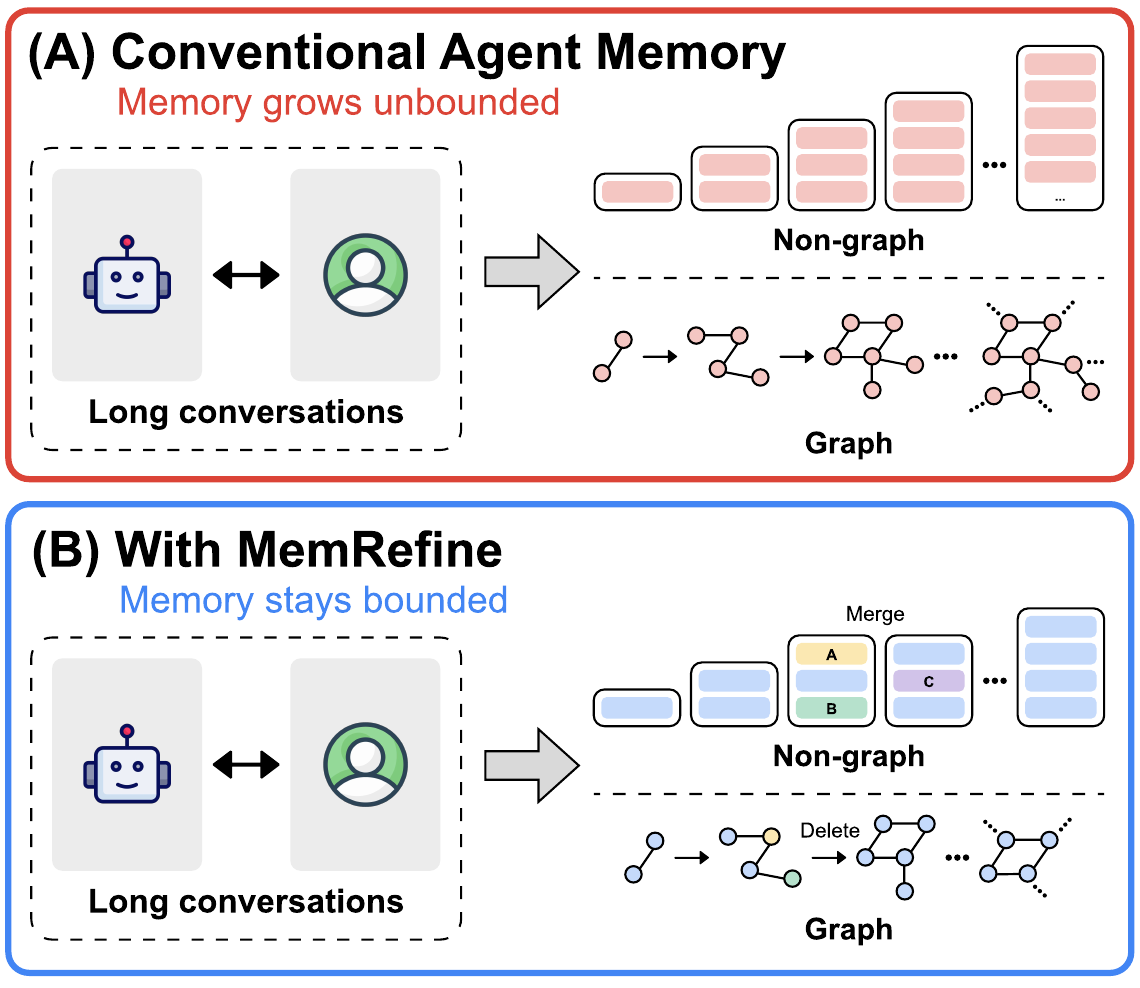}
\vspace{-0.05in}
\caption{
Conceptual overview of storage-budgeted memory management.
As long-term interactions accumulate, agent memory grows without bound and fills with redundant entries, inflating storage cost.
\textsc{MemRefine} refines the already constructed store under a fixed storage budget, keeping it compact while preserving information useful for future retrieval.
}
\label{fig:concept}
\vspace{-0.075in}
\end{figure}

%% file: Sections/3_related_work.tex
\section{Related Work}

\paragraph{Long-Term Memory for LLM Agents}
Long-term memory has become a central component for dialogue systems and LLM agents that operate over information accumulated across extended interactions.
Early work on long-term and multi-session dialogue showed that conversational agents need to store, update, and retrieve information from previous sessions to maintain consistency over time \citep{xu2022beyond, bae2022keep, jang2023conversation}.
More recent work develops external memory systems for LLM agents, including user-level memory banks \citep{zhong2024memorybank}, explicit read-write memory modules \citep{modarressi2023retllm}, virtual context management \citep{packer2023memgpt}, structured note graphs \citep{xu2025amem}, and general-purpose memory layers for agents \citep{mem0}, alongside benchmarks that evaluate whether such systems can recall information over extended histories \citep{maharana2024locomo, wu2025longmemeval}.
However, these efforts predominantly focus on \emph{growing} memory, namely how new information is added and recalled as interactions accumulate, leaving the inverse problem of \emph{shrinking} an already constructed store under a storage budget largely unaddressed.
In contrast, we treat post-construction storage reduction as a standalone task, compressing an existing memory store to fit a target budget while leaving the host memory framework unchanged.

\paragraph{Memory and Context Compression}
Reducing the cost of storing or processing long-term information has been studied in related but distinct settings.
Session-level memory summarization compresses dialogue histories into compact summaries before they enter memory, for instance, through recursive summarization for long-term dialogue memory \citep{wang2023recursive} or compressive memory-enhanced dialogue generation \citep{chen2025comedy}.
In the meantime, inference-time context compression instead reduces the evidence or prompt fed to the LLM rather than the underlying store, including RECOMP for retrieved documents \citep{xu2024recomp} and LLMLingua and LongLLMLingua for long-context prompts \citep{jiang2023llmlingua, jiang2024longllmlingua}.
Structure-based reduction shrinks memory graphs offline via criteria such as centrality or connectivity \citep{brin1998anatomy, wang2026streammeco}, while recent agent-memory frameworks improve memory efficiency through filtering, consolidation, or lightweight updates interleaved with the construction pipeline \citep{fang2025lightmem, liu2026simplemem}.

None of these approaches, however, takes \emph{reducing an already constructed memory store to a fixed storage budget through reasoning about the factual content of its entries} as the explicit objective: summarization preempts memory rather than shrinking it, context compression touches the inference-time prompt rather than the store, and structure- or update-based methods either rely on criteria insensitive to content or remain coupled to the construction pipeline.
In contrast, we fill this gap by directly compressing an existing memory store toward a target storage budget, applying LLM-guided judgments over semantically related entries to decide whether each pair carries redundant facts to be removed, complementary facts to be merged, or distinct facts to be preserved, while remaining decoupled from the underlying memory construction and retrieval pipelines.

%% file: Sections/4_method.tex
\input{Figures/method}

\section{Method}

We present \textsc{MemRefine}, an LLM-guided framework that compresses an already constructed agent memory store to a target storage budget while preserving information useful for future interactions.
We first formalize storage-budgeted memory management as a budget-constrained max-min program over query-agnostic memory compression, and introduce \textsc{MemRefine}, an LLM-guided pairwise refinement procedure that operationalizes it.

\subsection{Problem Formulation}
\label{sec:problem_formulation}

\paragraph{Notation.}
Let $M_0 = \{m_i\}_{i=1}^{n}$ denote an already constructed memory store, where each entry $m_i$ pairs textual content with an embedding vector and, depending on the host memory framework, may additionally carry links, timestamps, metadata, or other structured fields.
We assume a downstream pipeline that, given a query $q$ and a memory store $M$, retrieves a context $c = \texttt{Ret}(q; M)$ and produces an answer $a = \texttt{LLM}(q, c)$.
Downstream utility is then measured by a task-specific score $U(q, M) \in [0,1]$ that compares $a$ against a reference answer (e.g., token-level F1 on factual recall).

\paragraph{Storage-Budgeted Memory Management.}
We formalize storage-budgeted memory management as the task of producing a compressed memory store $M'$ from $M_0$ that fits within a fixed storage budget while remaining as useful as possible across an unknown future query distribution over a query space $\gQ$.
Formally, let $\gF(M_0)$ denote the set of memory stores reachable from $M_0$ through edit operations, such as deletions and merges of entries, and let $\rho \in (0,1]$ specify a storage budget ratio.
Since the compressor cannot tailor $M'$ to any particular query, we cast compression as a query-agnostic max-min program that hedges against the worst-case downstream query, defined as follows:
\begin{equation}
\label{eq:maxmin}
\begin{aligned}
    & \max_{M' \in \gF(M_0)} \;\; \min_{q \in \gQ} \;\; U(q, M') \\
    & \text{s.t.} \quad \mathrm{size}(M') \leq \rho \cdot \mathrm{size}(M_0),
\end{aligned}
\end{equation}
where the inner minimization picks out the worst-case query under $M'$, so that the outer maximization is forced to retain enough evidence for every plausible $q \in \gQ$ rather than only the easy ones\footnote{Equivalently, the dual form $\min_{M' \in \gF(M_0)} \mathrm{size}(M')$ subject to $U(q, M') \geq U(q, M_0) - \eps$ for all $q \in \gQ$ expresses the same intent of reducing memory size while uniformly preserving downstream utility within a tolerance $\eps$.}.

\paragraph{Query-Agnostic Compression.}
We note that directly optimizing Equation~\ref{eq:maxmin} is intractable: $\gQ$ is unobserved at compression time, $U$ requires running the full downstream pipeline for every candidate $M'$, and $\gF(M_0)$ is combinatorial in $n$.
Yet the inner minimization itself implies a useful constraint on removable content: since the compressor should hedge against every plausible query, it can only safely discard entries that are either \emph{redundant} (already covered by some other entry in $M'$) or \emph{non-factual} (carrying no information useful to any plausible query).
This contrasts with inference-time prompt or context compressors~\citep{jiang2023llmlingua, xu2024recomp, jiang2024longllmlingua}, which use the incoming query as a selection signal to prune evidence at the token level.
In contrast, we treat compression as a \emph{preservation} problem and operationalize this view by replacing the intractable inner minimization with a tractable per-pair surrogate, namely a judge (powered by an LLM) that, for each pair of semantically related entries, decides whether their content is redundant, complementary, or distinct, and authorizes only those edits that preserve facts not redundantly stored elsewhere.

\subsection{\textsc{MemRefine}}
\label{sec:memrefine}

\paragraph{Pairwise Refinement.}
A direct alternative to the per-pair surrogate above would be to assign an importance score to each memory entry independently and remove low-scoring entries until the budget is met.
However, this fails to capture the central property that motivates Equation~\ref{eq:maxmin}: the value of a memory entry is \emph{relational} rather than independent, since an entry may appear marginal in isolation yet become indispensable when combined with another; conversely, an entry may look high-value yet be safely removable because the same fact is already stored elsewhere in $M'$.
These two failure modes look identical under any per-entry criterion (e.g., embedding norm, recency, or graph centrality), so single-entry scoring cannot tell when an edit removes a fact not redundantly covered elsewhere.
To resolve this, \textsc{MemRefine} operates on \emph{pairs} of semantically related memory entries: for any pair $(u, v)$, the joint content decomposes exhaustively into a redundant case (the content of one is already covered by the other), a complementary case (each entry contributes content the other lacks), and a distinct case (the two entries assert unrelated facts), which directly induces a \emph{delete}--\emph{merge}--\emph{preserve} action space matching the edits admitted by $\gF(M_0)$.
Furthermore, \textsc{MemRefine} realizes this by iterating over candidate pairs and selecting one action per pair until the budget is met or no unresolved candidate remains; Algorithm~\ref{alg:memrefine} summarizes the procedure, with each step detailed below.

\begin{algorithm}[t]
\small
\caption{\textsc{MemRefine} compression procedure.}
\label{alg:memrefine}
\begin{algorithmic}[1]
\Require Input memory store $M_0$, storage budget ratio $\rho$
\Ensure Compressed memory store $M'$

\State $B \gets \rho \cdot \mathrm{size}(M_0)$
\State $M' \gets M_0$
\State $\mathcal{S} \gets \emptyset$ \Comment{Pairs judged as \textsc{preserve}}

\While{$\mathrm{size}(M') > B$}
    \State $(u,v) \gets \texttt{MostSimilarPair}(M', \mathcal{S})$
    \If{$(u,v) = \emptyset$}
        \State \textbf{break}
    \EndIf

    \State $d \gets \texttt{Judge}(u,v)$

    \If{$d.\mathrm{action} = \textsc{delete}$}
        \State Remove $d.\mathrm{target}$ from $M'$
        \State Prune dangling links if links exist

    \ElsIf{$d.\mathrm{action} = \textsc{merge}$}
        \State $m \gets \texttt{Merge}(u, v, d.\mathrm{instruction})$
        \State Replace $u$ and $v$ with $m$ in $M'$
        \State Inherit the union of links if links exist
        \State Recompute embedding for $m$

    \Else
        \State $\mathcal{S} \gets \mathcal{S} \cup \{(u,v)\}$
    \EndIf
\EndWhile

\State \Return $M'$
\end{algorithmic}
\end{algorithm}

\paragraph{Candidate Pair Selection.}
At each iteration, \textsc{MemRefine} selects the most similar pair of entries that has not yet been judged, denoted as follows: $(u, v) \gets \arg\max_{(u, v) \notin \mathcal{S}} \cos(e_u, e_v)$, where $e_u$ denotes the embedding of $m_u$ and $\mathcal{S}$ caches pairs previously judged as \textsc{preserve} to avoid revisiting them in subsequent iterations.
Notably, similarity here serves only as a \emph{proposal} mechanism: it surfaces pairs whose content is most likely to overlap, but does not determine whether either entry should be removed.
This separation matters because high cosine similarity can equally indicate redundancy, complementarity, or merely topical overlap, all of which call for different actions.

\paragraph{LLM-Guided Action.}
The selected pair $(u, v)$ is then passed to an LLM judge: $d \gets \texttt{Judge}(u, v)$, which returns a structured decision $d$ containing an action $d.\mathrm{action} \in \{\textsc{delete}, \textsc{merge}, \textsc{preserve}\}$, an optional deletion target, and (when the action is \textsc{merge}) a merge instruction specifying which facts should be retained.
The judge is prompted to reason about the factual content of $u$ and $v$, so that lexically similar entries with distinct facts are not collapsed, and lexically dissimilar entries that nonetheless encode the same fact are recognized as redundant (See Appendix~\ref{app:prompts} for the full prompt).

\paragraph{Store Update.}
Given the returned action, \textsc{MemRefine} triggers a corresponding update to $M'$:
\begin{itemize}[itemsep=0.0mm, parsep=1pt, leftmargin=*]
    \item \textsc{delete}: the judge-selected target is removed from $M'$, and dangling edges incident to it are pruned if the memory framework contains links;
    \item \textsc{merge}: a separate merge LLM, conditioned on the judge instruction, synthesizes a single entry $m \gets \texttt{Merge}(u, v, d.\mathrm{instruction})$ that replaces both $u$ and $v$ in $M'$, inheriting the union of their links (excluding duplicates and self-loops) along with a freshly recomputed embedding so that subsequent proposals reflect the merged content;
    \item \textsc{preserve}: both entries remain in $M'$ unchanged, and $(u, v)$ is added to $\mathcal{S}$.
\end{itemize}

\paragraph{Termination.}
The loop continues until either the budget $\mathrm{size}(M') \leq B$ is met or no unjudged candidate pair remains.
This early-exit behavior reflects the design principle that factual preservation takes precedence over budget compliance: when all sufficiently similar pairs have already been judged \textsc{preserve}, further compression would necessarily remove facts not redundantly stored elsewhere, violating the inner minimization in Equation~\ref{eq:maxmin}.

\paragraph{Practical Implications.}
It is worth noting that the proposed \textsc{MemRefine} is designed as an offline, post-construction module that sits between memory construction and retrieval: all judge and merge calls are issued during memory maintenance rather than at query time, so compression reduces the persistent footprint of the store without inflating per-query retrieval or generation latency.
It is also agnostic to the host memory representation, maintaining the link structure (pruning dangling edges on \textsc{delete}, inheriting the union of links on \textsc{merge}) for graph-structured frameworks~\citep{xu2025amem} and operating directly on entries and their embeddings for non-graph stores~\citep{mem0}; in both cases, the host construction and retrieval pipelines remain untouched.

%% file: Figures/method.tex
\begin{figure*}[t]
\centering
\includegraphics[width=\textwidth]{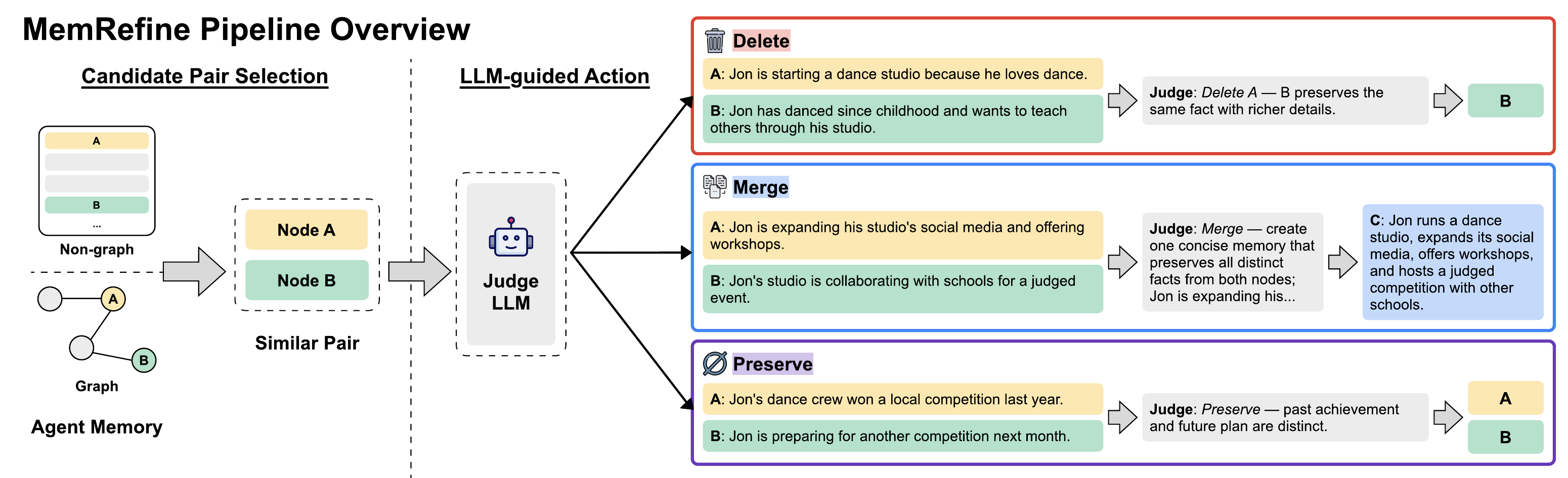}
\vspace{-0.25in}
\caption{
Overview of the \textsc{MemRefine} pipeline.
Given an already constructed memory store, \textsc{MemRefine} proposes semantically similar memory pairs and lets an LLM judge delete redundant, merge complementary, or preserve distinct memories, iterating until the store meets the storage budget.
}
\label{fig:method}
\vspace{-0.05in}
\end{figure*}

%% file: Sections/5_experimental_setup.tex
\section{Experimental Setup}

Our evaluation tests whether an already constructed long-term memory store can be compressed under an explicit storage budget while preserving downstream utility.
To stress this claim, we vary three factors that directly govern how much of the store can be removed without degrading downstream utility: the host memory framework, the compression budget, and the length of the interaction history.

\paragraph{Datasets.}
We evaluate our \textsc{MemRefine} on long-term conversational memory benchmarks, whose questions require recalling evidence dispersed across extended, multi-session dialogue histories.
Our primary benchmark is \textbf{LoCoMo}~\citep{maharana2024locomo}, for which we use 10 samples comprising 1,986 questions spanning five categories, namely single-hop factual recall, multi-hop reasoning, temporal reasoning, open-ended, and counterfactual questions, which together probe whether compression preserves both localized facts and evidence spread across the store.
To further stress-test compression under longer and more redundant histories, we construct \textbf{scaled LoCoMo-style} datasets with roughly 3x and 10x longer conversations, which amplify topic recurrence and redundancy while also raising the risk that rare but useful facts are removed under aggressive compression (full construction and quality statistics are reported in Appendix~\ref{app:scaled_quality}).
Finally, to verify that our findings are not specific to LoCoMo-style evaluation, we additionally evaluate on \textbf{LongMemEval}$_S$~\citep{wu2025longmemeval}, using 60 questions balanced across its question types with 10 questions per type.

\paragraph{Memory frameworks.}
To examine whether post-construction compression generalizes across different memory representations, we apply \textsc{MemRefine} to two representative frameworks that store memory in markedly different forms.
In particular, \textbf{A-MEM}~\citep{xu2025amem} maintains a graph of linked structured notes; following its design, we adopt an A-MEM-style factual graph representation that retains the memory content, embeddings, and links while excluding the auxiliary context and tag fields, which isolates compression over factual graph memory (a supporting field analysis is provided in Appendix~\ref{app:amem_representation}).
\textbf{Mem0}~\citep{mem0}, in contrast, constructs entries through an ingestion-and-update pipeline that transforms raw dialogue into processed memory rather than preserving it as raw dialogue nodes.
In both settings, \textsc{MemRefine} is inserted after memory construction and before retrieval, leaving the host construction and retrieval pipelines unchanged.

\input{Tables/main_benchmark_results}

\paragraph{Compression budgets and baselines.}
For each framework, we apply \textsc{MemRefine} under a range of storage budgets, namely 70\%, 60\%, 50\%, 40\%, and 30\% of the uncompressed store, which lets us trace the full storage-utility trade-off from mild to aggressive compression.
The primary point of comparison is the corresponding uncompressed \textbf{Base Memory}, against which we measure how much downstream utility is retained at each budget.
To isolate the contribution of the LLM judge at the core of our method, we further compare against two rule-based baselines on A-MEM-style graph memory that reuse the same candidate-pair selection loop but replace factual judgment with fixed heuristics: \textsc{RuleSim}, which decides each action from a fixed embedding-similarity threshold, and \textsc{RulePR}, which leverages an additional graph representation through PageRank-style preservation.
These baselines directly test whether surface similarity or graph centrality alone is sufficient to decide which facts should be deleted, merged, or preserved.
We restrict this rule-based comparison to graph memory because both baselines depend on graph structure; further details are in Appendix~\ref{app:rule_based_baselines}.

\paragraph{Evaluation.}
On LoCoMo, we report token-level F1 and exact match (EM) following the LoCoMo evaluation protocol~\citep{maharana2024locomo}, and additionally break F1 down by question category to examine how compression affects different reasoning types.
On LongMemEval$_S$, we report accuracy, with GPT-4o-mini serving as the automatic judge.
Alongside downstream utility, we track the realized storage ratio of each compressed store to confirm that it meets the target budget.

\paragraph{Implementation details.}
For candidate-pair selection, we embed each entry in the memory with a \texttt{all-MiniLM-L6-v2} model and rank pairs by cosine similarity, while the downstream QA uses the default retriever of each host framework with top-$k=10$.
Unless otherwise noted, both the judge and merge models are \texttt{gpt-5-mini}; for the compression model analysis, we also evaluate \texttt{Qwen3-8B} as the judge and merge model while keeping the retrieval and QA pipeline fixed.
Additional implementation details are provided in Appendix~\ref{app:impl}, rule-based baseline details in Appendix~\ref{app:rule_based_baselines}, threshold selection in Appendix~\ref{app:threshold_sensitivity}, prompts in Appendix~\ref{app:prompts}, and full model-family results in Appendix~\ref{app:model_family}.

%% file: Tables/main_benchmark_results.tex
\providecolor{memgray}{HTML}{F2F2F2}

\begin{table*}[t]
\centering
\small
\setlength{\dashlinedash}{0.8pt}
\setlength{\dashlinegap}{1.8pt}
\renewcommand{\arraystretch}{0.975}
\setlength{\tabcolsep}{6pt}
\begin{tabular*}{\textwidth}{@{\extracolsep{\fill}} l c c c c c c c}
\toprule
 & & & \multicolumn{5}{c}{\textbf{Storage Budget (\%)}} \\
\cmidrule(lr){4-8}
\textbf{Framework} & \textbf{Metric} & \textbf{Base} & \textbf{70} & \textbf{60} & \textbf{50} & \textbf{40} & \textbf{30} \\
\midrule

\rowcolor{memgray}
\multicolumn{8}{c}{\rule{0pt}{2.0ex}\textbf{\textit{LoCoMo}}} \\
\midrule
\multirow{2}{*}{A-MEM graph} & F1
& 0.4013 & \textbf{0.4014} & 0.3977 & 0.3902 & 0.3844 & 0.3628 \\
            & EM
& 0.1712 & 0.1690 & \textbf{0.1715} & 0.1628 & 0.1615 & 0.1474 \\
\cdashline{1-8}
\multirow{2}{*}{Mem0}        & F1
& 0.2827 & \textbf{0.2888} & 0.2773 & 0.2761 & 0.2685 & 0.2510 \\
            & EM
& 0.1098 & \textbf{0.1177} & 0.1095 & 0.1132 & 0.1102 & 0.0994 \\

\midrule
\rowcolor{memgray}
\multicolumn{8}{c}{\rule{0pt}{2.0ex}\textbf{\textit{3x LoCoMo}}} \\
\midrule
\multirow{2}{*}{A-MEM graph} & F1
& 0.2344 & 0.2307 & \textbf{0.2324} & 0.2163 & 0.2079 & 0.1900 \\
            & EM
& 0.0890 & 0.0893 & \textbf{0.0923} & 0.0879 & 0.0779 & 0.0702 \\
\cdashline{1-8}
\multirow{2}{*}{Mem0}        & F1
& 0.2204 & 0.2152 & 0.2065 & \textbf{0.2085} & 0.2013 & 0.1900 \\
            & EM
& 0.0588 & 0.0643 & 0.0621 & \textbf{0.0651} & 0.0599 & 0.0559 \\

\midrule
\rowcolor{memgray}
\multicolumn{8}{c}{\rule{0pt}{2.0ex}\textbf{\textit{10x LoCoMo}}} \\
\midrule
\multirow{2}{*}{A-MEM graph} & F1
& 0.2053 & \textbf{0.2033} & 0.2006 & 0.1916 & 0.1776 & 0.1660 \\
            & EM
& 0.0786 & 0.0795 & \textbf{0.0798} & 0.0723 & 0.0659 & 0.0574 \\
\cdashline{1-8}
\multirow{2}{*}{Mem0}        & F1
& 0.2039 & \textbf{0.1986} & 0.1955 & 0.1899 & 0.1858 & 0.1800 \\
            & EM
& 0.0634 & \textbf{0.0617} & 0.0603 & 0.0600 & 0.0581 & 0.0564 \\

\midrule
\rowcolor{memgray}
\multicolumn{8}{c}{\rule{0pt}{2.0ex}\textbf{\textit{LongMemEval$_S$}}} \\
\midrule
A-MEM graph & Acc.
& 0.5833 & 0.6000 & 0.6000 & \textbf{0.6167} & 0.5833 & 0.5500 \\
\cdashline{1-8}
Mem0        & Acc.
& 0.5167 & 0.4833 & \textbf{0.5167} & 0.4833 & 0.4500 & 0.4000 \\

\bottomrule
\end{tabular*}
\vspace{-0.075in}
\caption{
Main storage-performance results across benchmarks, scaled conversation histories, and memory frameworks.
\textbf{Bold} marks the strongest compressed result.
Per-category results, LongMemEval$_S$ type-level results, and full scaled results with node counts and storage sizes are reported in Appendix~\ref{app:category_results}, \ref{app:longmemeval}, and \ref{app:scaled_full}, respectively.
}
\label{tab:main_benchmark_results}
\vspace{-0.1in}
\end{table*}

%% file: Sections/6_experimental_results.tex
\input{Figures/category_delta_heatmap_main}
\input{Figures/rule_baseline_tradeoff}

\section{Experimental Results and Analysis}
\label{sec:results}

We structure our evaluation around three questions: whether an already constructed memory store can be compressed without sacrificing downstream utility, whether LLM-guided factual judgment is necessary or similarity and graph structure already suffice, and whether the storage-utility trade-off remains favorable as conversations grow longer.

\paragraph{Overall Results.}
Table~\ref{tab:main_benchmark_results} reports how the storage budget affects downstream performance across the standard LoCoMo, scaled LoCoMo-style setups, and LongMemEval$_S$.
The central observation is that \textsc{MemRefine} substantially reduces memory size while preserving downstream utility, exposing a controllable storage-utility trade-off that holds across all settings.
On standard LoCoMo, both A-MEM-style graph memory and Mem0 retain performance comparable to the uncompressed store under moderate compression, with F1 holding essentially flat even as a large share of entries is removed, suggesting that constructed memory stores carry redundant or fragmented entries that can be removed or consolidated without losing answer-bearing content.
The scaled LoCoMo-style settings provide a harder test, since longer conversations introduce more redundancy while also raising the chance of touching rare but useful facts; even here, \textsc{MemRefine} preserves utility smoothly across budgets, indicating that it remains effective as interaction histories grow.
On LongMemEval$_S$, A-MEM-style graph memory again benefits from moderate compression and even reaches its best accuracy at an intermediate budget, while Mem0 preserves utility across a narrower band of budgets.
The removable headroom thus depends on the host representation, larger for A-MEM-style graph memory than for the already-consolidated Mem0, yet \textsc{MemRefine} compresses both unchanged as a general post-construction layer rather than framework-specific.

\input{Figures/model_family_tradeoff_main}

\paragraph{Category-Wise Effects.}
The benchmark-level results show that moderate compression preserves overall performance, but they do not reveal how this effect is distributed across question types.
To examine this, Figure~\ref{fig:category_delta_main} reports category-wise F1 changes on LoCoMo relative to the uncompressed Base Memory of each framework (with full numerical results in Appendix~\ref{app:category_results}). As shown in Figure~\ref{fig:category_delta_main}, the effect of compression varies across categories.
In A-MEM-style graph memory, localized categories such as single-hop and temporal questions tend to improve after compression, consistent with the intuition behind post-construction refinement: removing or consolidating repeated facts makes retrieval less noisy when the answer depends on a small, localized piece of evidence.
Open-ended and counterfactual questions benefit less and are more sensitive to tighter budgets, since their evidence is spread across a wider portion of the store.
Mem0 exhibits a related but distinct category profile, reflecting that its ingestion-and-update pipeline already transforms raw interactions into processed entries before \textsc{MemRefine} is applied.
Together, these results show that the effect of compression is controlled jointly by the question type and memory representation produced by the host framework.

\paragraph{Comparison with Rule-based Graph-Memory.}
To isolate the role of the LLM judge, we compare \textsc{MemRefine} against rule-based baselines that retain the same candidate-pair selection but replace factual judgment with fixed heuristics, directly testing our central design choice that similarity is useful for proposing candidate pairs yet insufficient for deciding the compression action.
Figure~\ref{fig:rule_baseline_tradeoff} shows that rule-based methods remain competitive when the budget is loose, where most decisions involve obvious near-duplicates, but they fall off far more sharply as the budget tightens.
This is expected: under aggressive compression the difficult cases are no longer near-duplicate nodes but semantically related memories whose factual relationship should be interpreted, and a fixed similarity threshold cannot separate redundant paraphrases from complementary facts, while graph centrality does not indicate whether a memory holds evidence needed for future questions.
The widening advantage of \textsc{MemRefine} under tighter budgets therefore supports the need for LLM-guided factual action selection.

\paragraph{Effect of Compression Model.}
Since \textsc{MemRefine} depends on an LLM judge and merge model, we next ask whether its behavior is tied to a single model family.
Figure~\ref{fig:model_family_main} compares GPT-5-mini with the open Qwen3-8B while holding the downstream retrieval and QA pipeline (with full numerical results in Appendix~\ref{app:model_family}).
The two models behave similarly under moderate compression, indicating that \textsc{MemRefine} does not rely on a single proprietary judge; the gap widens under tighter budgets, where compression requires finer distinctions between redundant, complementary, and distinct memories.
Stronger judge models therefore become more valuable as the storage constraint grows severe.

\input{Figures/scaled_tradeoff}

\paragraph{Scaling to Longer Conversations.}
Finally, we ask whether compression remains well-behaved once histories grow much longer.
Figure~\ref{fig:scaled_tradeoff} plots the storage-utility curve on the scaled LoCoMo-style datasets, and the trade-off stays smooth and progressive across both the 3x and 10x settings.
This smooth, predictable behavior (rather than an abrupt failure point) suggests that \textsc{MemRefine} can serve as a practical maintenance mechanism for long-running agents whose stores keep growing.

%% file: Figures/category_delta_heatmap_main.tex
\begin{figure*}[t]
\centering
\includegraphics[width=0.92\textwidth]{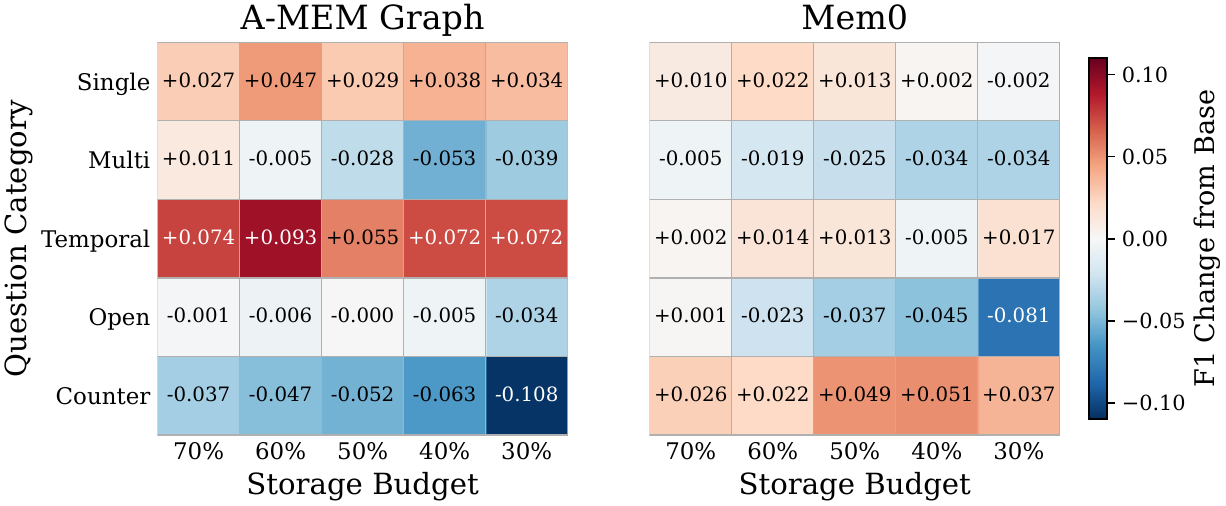}
\caption{
Category-wise F1 change under \textsc{MemRefine} compression on LoCoMo, measured against the uncompressed Base Memory.
Positive values indicate improvement.
}
\label{fig:category_delta_main}
\vspace{-0.05in}
\end{figure*}

%% file: Figures/rule_baseline_tradeoff.tex
\begin{figure}[t]
\centering
\includegraphics[width=0.90\columnwidth]{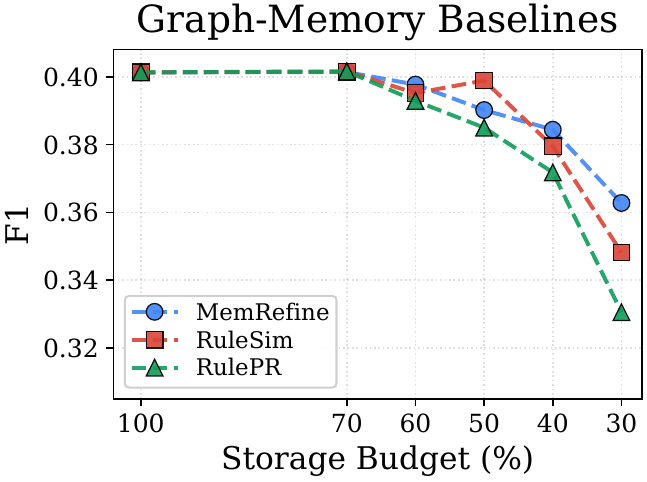}
\caption{
Comparison of \textsc{MemRefine} with rule-based baselines (\textsc{RuleSim} and \textsc{RulePR}) on A-MEM-style graph memory across storage budgets.
}
\label{fig:rule_baseline_tradeoff}
\vspace{-0.05in}
\end{figure}

%% file: Figures/model_family_tradeoff_main.tex
\begin{figure*}[t]
\centering
\includegraphics[width=0.82\textwidth]{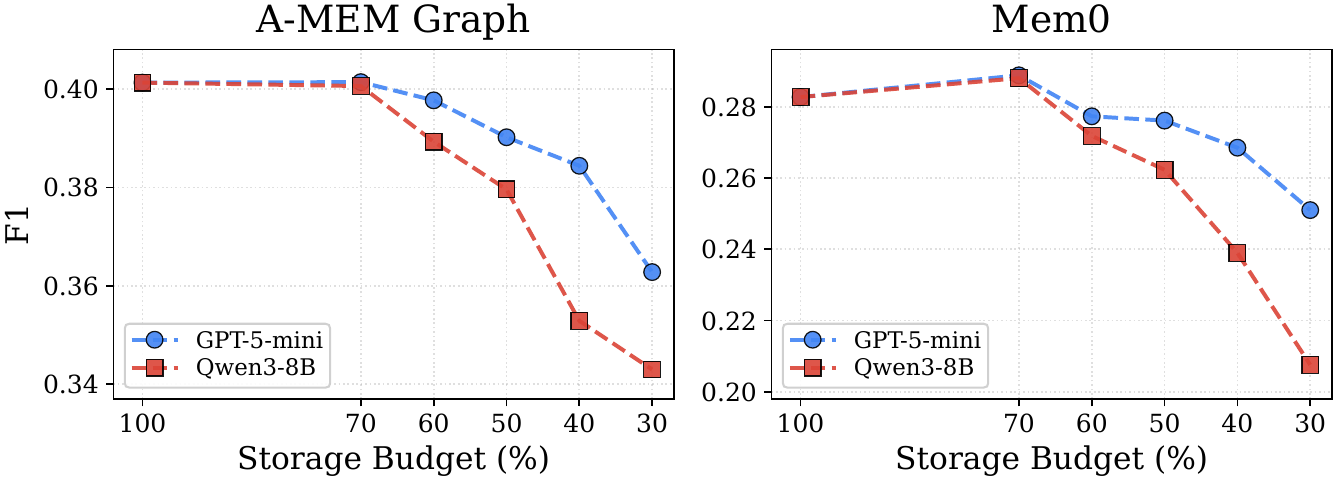}
\vspace{-0.05in}
\caption{
Effect of the compression model: \textsc{MemRefine} with GPT-5-mini versus the open Qwen3-8B as judge and merge model across storage budgets.
}
\label{fig:model_family_main}
\vspace{-0.05in}
\end{figure*}

%% file: Figures/scaled_tradeoff.tex
\begin{figure}[t]
\centering
\includegraphics[width=0.975\columnwidth]{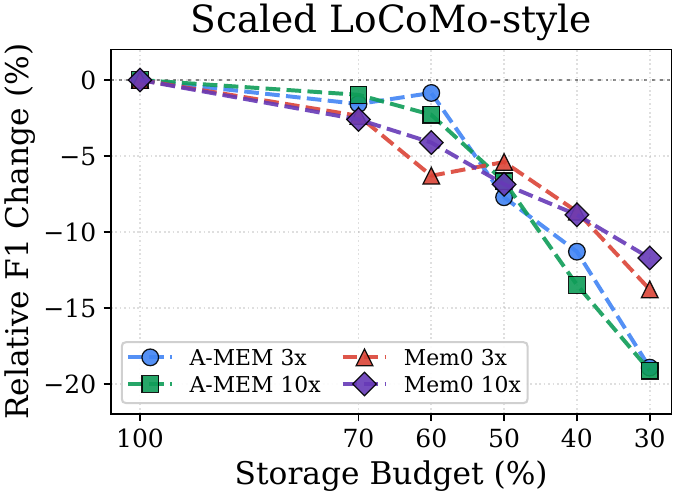}
\vspace{-0.05in}
\caption{
Storage-utility trade-off of \textsc{MemRefine} on the scaled LoCoMo-style datasets, with 3x and 10x longer conversations than standard LoCoMo.
}
\label{fig:scaled_tradeoff}
\vspace{-0.05in}
\end{figure}

%% file: Sections/7_conclusion.tex
\section{Conclusion}

In this paper, we proposed \textsc{MemRefine}, a post-construction memory compression framework designed to keep an already constructed long-term agent memory store within a fixed storage budget while preserving information useful for future interactions. Through a pairwise refinement procedure that uses similarity only to surface candidate pairs and defers the delete, merge, and preserve decision to an LLM judge grounded in factual content, \textsc{MemRefine} can be inserted between memory construction and retrieval without modifying the host memory construction pipeline. Empirical evaluations across A-MEM-style graph memory, Mem0, scaled LoCoMo-style settings, and LongMemEval$_S$ demonstrate that moderate compression preserves downstream utility while tighter budgets expose a gradual storage-utility trade-off, with the gap over rule-based baselines widening precisely as the budget tightens. Also, our analyses highlight that the amount of safely removable content is shaped not only by the compressor but also by the host memory representation, with A-MEM-style graph memory exposing more removable redundancy and Mem0, having already filtered and consolidated entries during ingestion, leaving less room before utility is affected. We believe these findings establish post-construction compression as a practical maintenance layer for storage-budgeted long-term agent memory, suggesting that compression and construction should be considered jointly rather than as independent stages.

%% file: Sections/8_Limitation.tex
\section*{Limitations}

The proposed \textsc{MemRefine} is designed as a post-construction module that compresses an already constructed long-term memory store to a target storage budget while preserving information useful for downstream tasks. It is worth noting that our evaluation centers on two representative memory frameworks (an A-MEM-style graph memory of linked structured notes and the Mem0 ingestion-and-update pipeline) and on long-term conversational benchmarks, which together cover both graph-structured and processed memory stores but do not exhaust the space of agent memory systems; extending post-construction compression to additional memory architectures and deployed multi-agent or tool-use settings would be a valuable direction for future work. Also, our scaled LoCoMo-style benchmarks serve as controlled stress tests for studying how compression behaves as conversation length and redundancy grow, and evaluating \textsc{MemRefine} on real longitudinal memory logs collected over extended deployments would be an exciting next step for its practical applicability.

%% file: Sections/9_ethical_considerations.tex
\section*{Ethical Considerations}

The proposed \textsc{MemRefine} edits an already constructed long-term memory store through \textsc{delete} and \textsc{merge} operations, which persistently remove or consolidate past interactions and, in deployed systems, may involve sensitive user information. While our experiments use public research benchmarks and synthetically generated scaled LoCoMo-style datasets without collecting or releasing any real user data, in practice, compression should be applied under appropriate privacy safeguards, access control, and data-retention policies, and we further recommend pairing it with mechanisms such as soft deletion or audit logs of applied edits so that maintenance decisions remain reviewable.

%% file: Sections/10_appendix.tex
\clearpage
\appendix

\section{A-MEM Input Representation}
\label{app:amem_representation}

A-MEM represents each memory as a structured note with multiple fields, including content, contextual descriptions, keywords, tags, embeddings, and links.
For the graph-based memory setting, we use a simplified A-MEM-style representation that retains memory content, embeddings, and links while excluding auxiliary context and tag fields.
This design focuses the experiment on compressing factual memory entries and their graph structure, rather than evaluating A-MEM's memory evolution fields.
All A-MEM-based compression methods therefore start from the same Base Memory representation.

\input{Tables/appendix_amem_representation}

This field analysis suggests that the auxiliary context and tag fields do not provide useful evidence for the downstream retrieval setting used in our experiments, and may introduce additional retrieval noise.
We therefore use the simplified factual graph representation consistently across all A-MEM-style experiments.

\FloatBarrier

\subsection{Prompt Details}
\label{app:prompts}

\textsc{MemRefine} uses two LLM calls during compression: a judge LLM and a merge LLM.
For both calls, each memory content field is truncated to the first 2,000 characters before being inserted into the prompt.
The judge returns one of \texttt{delete}, \texttt{merge}, or \texttt{nothing}; the implementation maps \texttt{nothing} to the \textsc{Preserve} action described in the main paper.

\subsubsection{Judge Prompt}

\begin{PromptBox}
You are a memory compression assistant. Your goal is to reduce memory size while preserving information that is useful for answering future questions about the conversation.

Two memory nodes have cosine similarity {similarity}.

[Node A]
{content_a}

[Node B]
{content_b}

IMPORTANT -- Judge by FACTUAL INFORMATION VALUE, not by surface wording:
- Facts worth preserving: names, dates, events, locations, preferences, plans, opinions, personal experiences, relationships
- Low-value content: greetings ("Hey!", "Good to talk"), generic thanks ("Thanks!", "Appreciate it"), small talk, emotional filler without specific facts
- Two nodes that only differ in wording but carry the same facts -> delete one
- If both nodes are low-value (e.g. both are just greetings or generic thanks) -> delete one, don't merge

Choose one action:
1. "delete" -- One node is redundant OR both carry the same facts (even if worded differently). Delete the less informative one.
   -> Specify which to delete ("a" or "b"). Keep the one with more specific facts.
2. "merge" -- Both contain DISTINCT FACTS that would be lost if either were deleted. Combine into one.
   -> Write a merge instruction specifying which FACTS (not phrasings) from each node must be preserved.
3. "nothing" -- Both contain clearly distinct, important factual information. Keep both.

Bias toward "delete" when nodes only differ in wording. Only choose "merge" when there are genuinely different facts in each node.

Return JSON:
- action: "delete", "merge", or "nothing"
- target: "a" or "b" (which to delete; set "a" if not deleting)
- merge_instruction: instruction for merge LLM (set "" if not merging)
- reason: brief explanation
\end{PromptBox}

\subsubsection{Merge Prompt}

\begin{PromptBox}
You are merging two memory nodes into one. Follow the instruction below.

[Merge Instruction]
{instruction}

[Node A]
{content_a}

[Node B]
{content_b}

Rules:
- Preserve all FACTS (names, dates, events, locations, preferences, plans, experiences) from both nodes
- Drop generic phrases, greetings, and emotional filler that don't carry factual content
- Keep it concise -- shorter is better as long as no facts are lost
- Maintain speaker attribution (who said what) when it matters for the fact

Return JSON:
- content: the merged memory content
- keywords: list of keywords for the merged memory
\end{PromptBox}

\FloatBarrier

\section{Implementation Details}
\label{app:impl}

\paragraph{Memory construction and retrieval.}
Unless otherwise stated, we follow the original A-Mem memory construction pipeline and represent each memory as a graph node with its content and metadata.
For candidate pair selection, we embed each memory entry using \texttt{all-MiniLM-L6-v2} and compute pairwise cosine similarities over the resulting embeddings.
For question answering, we use the default retriever of the memory framework with top-$k=\texttt{10}$ retrieved memories.
To control the context length and cost of LLM-based decisions, we truncate each memory content field to the first 2{,}000 characters before passing it to the judge and merge models.

\paragraph{Memory size accounting.}
We measure memory size by serializing the stored memory entries into JSON with \texttt{ensure\_ascii=False}.
For our compressed variants, memory size is computed over the 8 fields used by our implementation.
The original A-Mem output additionally contains \texttt{context}, \texttt{category}, \texttt{tags}, and \texttt{retrieval\_count}, resulting in 12 serialized fields.
For ratio-based compression, the target memory budget for each sample is set to $r$ times the corresponding no-compression memory size, where $r \in \{0.3,0.4,0.5,0.6,0.7\}$.

\paragraph{Rule-based compression.}
For the rule-based delete-then-merge compressor, we first identify redundant memory pairs whose cosine similarity is at least a threshold $\tau$.
We evaluate $\tau \in \{0.70,0.80,0.90\}$ for ratio-based budgets and additionally use $\tau \in \{0.75,0.85\}$ for absolute-budget experiments.
In the PR-based setting, PageRank is computed on the memory graph and the higher-PageRank memory is preferentially preserved within redundant clusters.
We use PR protection percentile $0.9$ unless otherwise specified, which protects the top 10\% PageRank nodes from being selected as merge anchors.
If deletion alone does not satisfy the target memory budget, we fall back to the A-Mem-style merge operation, where low-PageRank anchor nodes and their neighbors are summarized into a single memory node.
All compression operations preserve the original timestamps associated with the affected memories.

\paragraph{LLM-judged compression.}
For the LLM-judged compressor, we remove the fixed similarity threshold and instead sort candidate memory pairs by cosine similarity in descending order.
For each pair, the judge model decides one of three actions: \texttt{delete}, \texttt{merge}, or \texttt{nothing}.
For \texttt{delete}, the judge also selects which memory should be removed.
For \texttt{merge}, the judge produces a merge instruction, and a separate merge model generates the merged memory following that instruction.
For \texttt{nothing}, the pair is skipped and is not queried again.
Compression stops when the target memory budget is reached or when no remaining pair receives a compressive action.
Unless otherwise noted, both the judge and merge models are \texttt{gpt-5-mini}.

\paragraph{Datasets.}
For LoCoMo, we evaluate on 10 samples, containing 1{,}986 questions across five categories: single-hop, multi-hop, temporal, open-ended, and counterfactual questions.
For LongMemEval$_S$, we evaluate on 60 questions across six types: single-user, single-assistant, single-preference, multi-session, temporal, and knowledge-update questions.
To stress-test longer histories, we construct scaled LoCoMo-style datasets at two scales (3x and 10x): the 3x setting comprises 5 long-dialogue samples (80 sessions and 1{,}600 user-assistant turns each, 2{,}720 QA pairs in total), and the 10x setting comprises 5 samples (270 sessions and 5{,}400 turns each, 9{,}180 QA pairs in total); full statistics are reported in Appendix~\ref{app:scaled_quality}.
For both scales, personas and events are generated with \texttt{gpt-5-mini}, while dialogues are generated with \texttt{gpt-4o-mini}.

\paragraph{Evaluation.}
For LoCoMo and the scaled LoCoMo-style datasets, we report token-level F1 and exact match (EM) following the LoCoMo evaluation protocol~\citep{maharana2024locomo}.
We also report category-wise F1 for the five LoCoMo question categories.
For LongMemEval$_S$, we report accuracy.
Unless otherwise noted, answer evaluation for the LLM-judged compression experiments uses \texttt{gpt-5-mini}.
For compression-model analysis, we use \texttt{Qwen3-8B}. All reported results are from a single compression and evaluation run for each setting due to the cost of API-based judge, merge, and evaluation calls; therefore, we do not report confidence intervals or error bars.

\section{Rule-based Graph-memory Baselines}
\label{app:rule_based_baselines}

We evaluate two rule-based baselines on A-MEM-style graph memory.
These baselines are used only in the graph-memory setting because they rely on graph structure or fixed similarity rules over graph nodes.

\subsection{\textsc{RuleSim}}

\textsc{RuleSim} uses the same pair-selection loop as \textsc{MemRefine} but replaces the LLM judge with a fixed similarity rule.
At each step, it selects the most similar unresolved memory pair $(u,v)$ with cosine similarity $s(u,v)$.
The compression action is determined as follows:
\[
\begin{cases}
\textsc{delete}, & s(u,v) \geq \tau_{\mathrm{delete}},\\
\textsc{merge}, & s(u,v) < \tau_{\mathrm{delete}}.
\end{cases}
\]
If the selected action is \textsc{delete}, one node is removed according to a deterministic deletion rule.
If the selected action is \textsc{merge}, the same merge LLM used by \textsc{MemRefine} generates the merged memory.
Thus, \textsc{RuleSim} isolates the effect of replacing the LLM judge with a fixed similarity decision rule.

\subsection{\textsc{RulePR}}

\textsc{RulePR} combines similarity-threshold deletion with PageRank-guided preservation and merge fallback.
It first identifies near-duplicate memory nodes whose cosine similarity exceeds a threshold.
Within each redundant group, PageRank is used as a graph-centrality proxy to decide which node should be preserved.
If deletion alone is insufficient to satisfy the storage budget, \textsc{RulePR} applies a merge fallback guided by the graph structure.

This baseline tests whether graph centrality is a useful substitute for LLM-guided factual compression decisions.
Unlike \textsc{MemRefine}, \textsc{RulePR} does not reason over factual redundancy or complementarity with an LLM judge.

\input{Tables/appendix_rule_baselines_full}
\input{Tables/appendix_threshold_sensitivity}

\FloatBarrier

\section{Threshold Selection for Rule-based Baselines}
\label{app:threshold_sensitivity}

For rule-based graph-memory baselines, we use $\tau = 0.8$ in the main experiments.
This value is selected based on threshold sweeps over the ratio-based graph-memory compression setting.
Table~\ref{tab:threshold_sensitivity} shows that $\tau = 0.8$ performs best at 70\%, 60\%, 50\%, and 40\% storage, while also providing a stable choice across storage budgets.
Although $\tau = 0.9$ is slightly better at 30\% storage, we use $\tau = 0.8$ as the main setting because it is the most consistently strong threshold across budgets.
\input{Tables/appendix_retention_only}

\FloatBarrier
\section{Retention-only Baselines}
\label{app:retention_only}

We also compare retention-only baselines that keep a fixed percentage of existing graph nodes without performing delete--merge compression.
The PageRank retention baseline keeps the top-ranked nodes according to graph centrality, while the random retention baseline keeps a random subset of nodes.
Table~\ref{tab:retention_only} shows that PageRank retention does not provide a reliable advantage over random retention.
Both retention-only baselines are substantially worse than compression-based methods, indicating that memory compression is not simply a node-selection problem.

\section{Category-wise LoCoMo Results}
\label{app:category_results}

The main paper visualizes category-wise F1 changes under compression on LoCoMo.
Here, Table~\ref{tab:category_full} reports the full numerical values for each question category and storage budget.

\section{Compression Model Analysis}
\label{app:model_family}

The main paper visualizes the effect of changing the judge and merge model.
Here, Table~\ref{tab:model_family_full} reports the full numerical results comparing GPT-5-mini with Qwen3-8B while keeping the downstream retrieval and QA pipeline fixed.
The results show that smaller open models can match the main compression pattern under moderate budgets, but stronger judge models are more robust when the storage budget becomes tight.

\FloatBarrier

\section[LongMemEval-S Full Results]{LongMemEval$_S$ Full Results}
\label{app:longmemeval}

The main paper summarizes LongMemEval$_S$ accuracy across storage budgets.
Table~\ref{tab:longmemeval_full} reports the full type-level results across all budgets.
LongMemEval$_S$ contains 60 questions, with 10 questions per type.
Accuracy is judged using GPT-4o-mini.

\FloatBarrier

\section{Full Scaled LoCoMo-style Results}
\label{app:scaled_full}

The main paper reports scaled LoCoMo-style results as a storage--performance curve.
Table~\ref{tab:scaled_full} provides the full numerical values for the 3x and 10x settings.

\FloatBarrier

\section{Scaled LoCoMo-style Dataset Details}
\label{app:scaled_quality}

We construct scaled LoCoMo-style datasets to evaluate memory compression under longer conversational histories.
Because the scaled datasets are synthetically generated, we verify that they preserve the broad structure of the original benchmark while increasing conversation length and redundancy.
Table~\ref{tab:scaled_dataset_quality} summarizes their basic statistics, text-level diversity indicators, and QA category distributions.
The 3x setting is a relatively faithful scaled version of LoCoMo, while the 10x setting serves as a stronger stress test with longer and more repetitive interaction histories.

\section{Artifact Licenses and Use}
\label{app:artifact_licenses}

We use publicly available research artifacts only for research and evaluation.
LoCoMo is released under a Creative Commons Attribution-NonCommercial license.
LongMemEval is distributed as a public research benchmark through its official repository and dataset release.
For memory frameworks, the A-MEM implementation is released under the MIT License, and Mem0 is released under the Apache License 2.0.
Our use of these artifacts is limited to benchmark evaluation and method comparison, and we do not redistribute the original benchmark data or framework code as part of this paper.
The scaled LoCoMo-style datasets are synthetic stress-test variants constructed for analysis of longer conversational histories.

\input{Tables/appendix_category_full}
\input{Tables/appendix_model_family_full}
\input{Tables/appendix_longmemeval_full}
\input{Tables/appendix_scaled_full}
\input{Tables/appendix_scaled_dataset_quality}

%% file: Tables/appendix_amem_representation.tex
\begin{table}[t]
\centering
\small
\setlength{\tabcolsep}{5pt}
\begin{tabular}{lrrr}
\toprule
Memory Representation & Size & F1 & EM \\
\midrule
Original A-MEM & 0.5430 & 0.3718 & 0.1385 \\
Base Memory & 0.2692 & 0.4013 & 0.1712 \\
\bottomrule
\end{tabular}
\caption{
Representative field-analysis result on LoCoMo.
Base Memory retains memory content, embeddings, and links while excluding auxiliary context and tag fields.
Size is reported in MB.
}
\label{tab:amem_representation}
\end{table}

%% file: Tables/appendix_rule_baselines_full.tex
\begin{table}[t]
\centering
\small
\setlength{\tabcolsep}{4pt}
\begin{tabular}{lrrrr}
\toprule
Method & Storage & Nodes & Size & F1 \\
\midrule
Base & 100\% & 588.2 & 0.2692 & 0.4013 \\
\textsc{MemRefine} & 70\% & 386.4 & 0.1888 & 0.4014 \\
 & 60\% & 318.9 & 0.1618 & 0.3977 \\
 & 50\% & 257.8 & 0.1349 & 0.3902 \\
 & 40\% & 197.5 & 0.1079 & 0.3844 \\
 & 30\% & 141.3 & 0.0811 & 0.3628 \\
\midrule
\textsc{RuleSim} & 70\% & -- & -- & 0.4015 \\
 & 60\% & -- & -- & 0.3952 \\
 & 50\% & -- & -- & 0.3989 \\
 & 40\% & -- & -- & 0.3795 \\
 & 30\% & -- & -- & 0.3482 \\
\midrule
\textsc{RulePR} & 70\% & 347.4 & 0.1882 & 0.4016 \\
 & 60\% & 268.1 & 0.1604 & 0.3929 \\
 & 50\% & 200.0 & 0.1340 & 0.3850 \\
 & 40\% & 135.9 & 0.1069 & 0.3718 \\
 & 30\% & 80.6 & 0.0793 & 0.3306 \\
\bottomrule
\end{tabular}
\caption{
Full graph-memory baseline comparison on LoCoMo.
Size is reported in MB when available.
}
\label{tab:rule_baselines_full}
\end{table}

%% file: Tables/appendix_threshold_sensitivity.tex
\begin{table}[t]
\centering
\small
\setlength{\tabcolsep}{6pt}
\begin{tabular}{lrrr}
\toprule
Storage & $\tau = 0.90$ & $\tau = 0.80$ & $\tau = 0.70$ \\
\midrule
70\% & 0.3880 & 0.4016 & 0.3763 \\
60\% & 0.3687 & 0.3929 & 0.3740 \\
50\% & 0.3634 & 0.3850 & 0.3828 \\
40\% & 0.3484 & 0.3718 & 0.3633 \\
30\% & 0.3347 & 0.3306 & 0.3330 \\
\bottomrule
\end{tabular}
\caption{
Threshold sensitivity for the PageRank-based rule baseline on A-MEM-style graph memory.
We report F1 under each storage budget.
}
\label{tab:threshold_sensitivity}
\end{table}

%% file: Tables/appendix_retention_only.tex
\begin{table}[t]
\centering
\small
\setlength{\tabcolsep}{4pt}
\begin{tabular}{lrrrr}
\toprule
Retention & Storage & Nodes & Size & F1 \\
\midrule
PageRank & 70\% & 417.5 & 0.1857 & 0.3111 \\
Random & 70\% & 411.4 & 0.1814 & 0.3085 \\
PageRank & 50\% & 294.5 & 0.1314 & 0.2673 \\
Random & 50\% & 293.7 & 0.1302 & 0.2848 \\
PageRank & 30\% & 176.8 & 0.0785 & 0.2098 \\
Random & 30\% & 175.9 & 0.0778 & 0.2374 \\
\bottomrule
\end{tabular}
\caption{
Retention-only baselines on A-MEM-style graph memory.
PageRank centrality is not a reliable criterion for selecting useful memories, and both retention-only baselines underperform compression-based methods.
Size is reported in MB.
}
\label{tab:retention_only}
\end{table}

%% file: Tables/appendix_category_full.tex
\begin{table*}[t]
\centering
\small
\setlength{\tabcolsep}{5pt}
\begin{tabular*}{\textwidth}{@{\extracolsep{\fill}}llrrrrr}
\toprule
Framework & Storage & Cat1 & Cat2 & Cat3 & Cat4 & Cat5 \\
\midrule
A-MEM graph & 100\% & 0.2961 & 0.4776 & 0.1754 & 0.4612 & 0.3310 \\
A-MEM graph & 70\% & 0.3235 & 0.4882 & 0.2498 & 0.4607 & 0.2938 \\
A-MEM graph & 60\% & 0.3434 & 0.4727 & 0.2687 & 0.4550 & 0.2842 \\
A-MEM graph & 50\% & 0.3251 & 0.4492 & 0.2303 & 0.4611 & 0.2789 \\
A-MEM graph & 40\% & 0.3345 & 0.4251 & 0.2477 & 0.4560 & 0.2681 \\
A-MEM graph & 30\% & 0.3300 & 0.4383 & 0.2470 & 0.4269 & 0.2226 \\
\midrule
Mem0 & 100\% & 0.2686 & 0.1246 & 0.1180 & 0.4266 & 0.1703 \\
Mem0 & 70\% & 0.2785 & 0.1194 & 0.1196 & 0.4276 & 0.1964 \\
Mem0 & 60\% & 0.2908 & 0.1060 & 0.1316 & 0.4036 & 0.1927 \\
Mem0 & 50\% & 0.2818 & 0.0996 & 0.1311 & 0.3897 & 0.2194 \\
Mem0 & 40\% & 0.2707 & 0.0904 & 0.1125 & 0.3812 & 0.2217 \\
Mem0 & 30\% & 0.2670 & 0.0908 & 0.1354 & 0.3456 & 0.2077 \\
\bottomrule
\end{tabular*}
\caption{
Category-wise F1 results on LoCoMo.
Cat1--Cat5 correspond to the LoCoMo question categories: single-hop, multi-hop, temporal, open-ended, and counterfactual questions.
}
\label{tab:category_full}
\end{table*}

%% file: Tables/appendix_model_family_full.tex
\begin{table*}[t]
\centering
\small
\setlength{\tabcolsep}{5pt}
\begin{tabular*}{\textwidth}{@{\extracolsep{\fill}}llrrrrrr}
\toprule
Framework & Model & 100\% & 70\% & 60\% & 50\% & 40\% & 30\% \\
\midrule
A-MEM graph & GPT-5-mini & 0.4013 & 0.4014 & 0.3977 & 0.3902 & 0.3844 & 0.3628 \\
A-MEM graph & Qwen3-8B & 0.4013 & 0.4006 & 0.3893 & 0.3796 & 0.3529 & 0.3430 \\
Mem0 & GPT-5-mini & 0.2827 & 0.2888 & 0.2773 & 0.2761 & 0.2685 & 0.2510 \\
Mem0 & Qwen3-8B & 0.2827 & 0.2880 & 0.2718 & 0.2623 & 0.2390 & 0.2076 \\
\bottomrule
\end{tabular*}
\caption{
Full compression-model comparison.
Values indicate F1 under each storage budget.
}
\label{tab:model_family_full}
\end{table*}

%% file: Tables/appendix_longmemeval_full.tex
\begin{table*}[t]
\centering
\scriptsize
\setlength{\tabcolsep}{3.5pt}
\resizebox{\textwidth}{!}{%
\begin{tabular}{llrrrrrrrr}
\toprule
System & Storage & Acc. & N & s-user & s-asst & s-pref & multi & temporal & k-update \\
\midrule
A-MEM graph & Base & 0.5833 & 60 & 9/10 & 10/10 & 3/10 & 4/10 & 3/10 & 6/10 \\
\midrule
A-MEM graph + \textsc{MemRefine} & 70\% & 0.6000 & 60 & 9/10 & 9/10 & 5/10 & 4/10 & 3/10 & 6/10 \\
A-MEM graph + \textsc{MemRefine} & 60\% & 0.6000 & 60 & 10/10 & 9/10 & 5/10 & 5/10 & 1/10 & 6/10 \\
A-MEM graph + \textsc{MemRefine} & 50\% & 0.6167 & 60 & 10/10 & 8/10 & 5/10 & 6/10 & 2/10 & 6/10 \\
A-MEM graph + \textsc{MemRefine} & 40\% & 0.5833 & 60 & 10/10 & 10/10 & 5/10 & 4/10 & 1/10 & 5/10 \\
A-MEM graph + \textsc{MemRefine} & 30\% & 0.5500 & 60 & 8/10 & 7/10 & 3/10 & 6/10 & 3/10 & 6/10 \\
\midrule
Mem0 & Base & 0.5167 & 60 & 8/10 & 4/10 & 6/10 & 5/10 & 4/10 & 4/10 \\
\midrule
Mem0 + \textsc{MemRefine} & 70\% & 0.4833 & 60 & 7/10 & 3/10 & 5/10 & 5/10 & 4/10 & 5/10 \\
Mem0 + \textsc{MemRefine} & 60\% & 0.5167 & 60 & 8/10 & 3/10 & 6/10 & 6/10 & 3/10 & 5/10 \\
Mem0 + \textsc{MemRefine} & 50\% & 0.4833 & 60 & 8/10 & 3/10 & 6/10 & 5/10 & 4/10 & 3/10 \\
Mem0 + \textsc{MemRefine} & 40\% & 0.4500 & 60 & 8/10 & 3/10 & 5/10 & 4/10 & 3/10 & 4/10 \\
Mem0 + \textsc{MemRefine} & 30\% & 0.4000 & 60 & 7/10 & 4/10 & 4/10 & 3/10 & 3/10 & 3/10 \\
\bottomrule
\end{tabular}%
}
\caption{
Full LongMemEval$_S$ results.
The benchmark contains 60 questions, with 10 questions per type.
Accuracy is judged using GPT-4o-mini.
}
\label{tab:longmemeval_full}
\end{table*}

%% file: Tables/appendix_scaled_full.tex
\begin{table*}[t]
\centering
\small
\setlength{\tabcolsep}{6pt}
\renewcommand{\arraystretch}{1.08}
\begin{tabular*}{\textwidth}{@{\extracolsep{\fill}}llrrrrr}
\toprule
Framework & Dataset & Storage & Nodes/Entries & F1 & EM & Rel. F1 Change \\
\midrule

\multirow{6}{*}{A-MEM graph}
& \multirow{6}{*}{3x}
& 100\% & 1600.0 & 0.2344 & 0.0890 & -- \\
& & 70\%  & 1038.2 & 0.2307 & 0.0893 & -1.6\% \\
& & 60\%  & 856.0  & 0.2324 & 0.0923 & -0.9\% \\
& & 50\%  & 701.6  & 0.2163 & 0.0879 & -7.7\% \\
& & 40\%  & 552.4  & 0.2079 & 0.0779 & -11.3\% \\
& & 30\%  & 398.6  & 0.1900 & 0.0702 & -18.9\% \\

\midrule

\multirow{6}{*}{A-MEM graph}
& \multirow{6}{*}{10x}
& 100\% & 5400.0 & 0.2053 & 0.0786 & -- \\
& & 70\%  & 3492.0 & 0.2033 & 0.0795 & -1.0\% \\
& & 60\%  & 2908.0 & 0.2006 & 0.0798 & -2.3\% \\
& & 50\%  & 2372.0 & 0.1916 & 0.0723 & -6.7\% \\
& & 40\%  & 1878.0 & 0.1776 & 0.0659 & -13.5\% \\
& & 30\%  & 1386.0 & 0.1660 & 0.0574 & -19.1\% \\

\midrule

\multirow{6}{*}{Mem0}
& \multirow{6}{*}{3x}
& 100\% & 1241.6 & 0.2204 & 0.0588 & -- \\
& & 70\%  & 868.6  & 0.2152 & 0.0643 & -2.4\% \\
& & 60\%  & 744.4  & 0.2065 & 0.0621 & -6.3\% \\
& & 50\%  & 620.6  & 0.2085 & 0.0651 & -5.4\% \\
& & 40\%  & 496.2  & 0.2013 & 0.0599 & -8.7\% \\
& & 30\%  & 372.0  & 0.1900 & 0.0559 & -13.8\% \\

\midrule

\multirow{6}{*}{Mem0}
& \multirow{6}{*}{10x}
& 100\% & 3531.0 & 0.2039 & 0.0634 & -- \\
& & 70\%  & 2471.4 & 0.1986 & 0.0617 & -2.6\% \\
& & 60\%  & 2118.2 & 0.1955 & 0.0603 & -4.1\% \\
& & 50\%  & 1765.2 & 0.1899 & 0.0600 & -6.9\% \\
& & 40\%  & 1412.0 & 0.1858 & 0.0581 & -8.9\% \\
& & 30\%  & 1058.6 & 0.1800 & 0.0564 & -11.7\% \\

\bottomrule
\end{tabular*}
\caption{
Full \textsc{MemRefine} results on scaled LoCoMo-style datasets.
The 3x and 10x settings contain approximately three and ten times longer conversations than standard LoCoMo, respectively.
Node or entry counts are reported as the average number of stored memory units after compression.
}
\label{tab:scaled_full}
\end{table*}

%% file: Tables/appendix_scaled_dataset_quality.tex
\begin{table*}[t]
\centering
\small
\setlength{\tabcolsep}{6pt}
\renewcommand{\arraystretch}{1.08}

\begin{tabular}{lrrrrr}
\toprule
\multicolumn{6}{c}{\textbf{(a) Basic statistics}} \\
\midrule
Dataset & Samples & Sessions & Turns & QAs & Avg. Turns/Sess. \\
\midrule
LoCoMo & 10 & 272   & 5,882  & 1,986 & 21.6 \\
3x     & 5  & 400   & 8,000  & 2,720 & 20.0 \\
10x    & 5  & 1,350 & 27,000 & 9,180 & 20.0 \\
\bottomrule
\end{tabular}

\vspace{0.75em}

\begin{tabular}{lrrr}
\toprule
\multicolumn{4}{c}{\textbf{(b) QA category distribution}} \\
\midrule
Category & LoCoMo & 3x & 10x \\
\midrule
Cat1 single-hop     & 14.2\% & 14.0\% & 14.0\% \\
Cat2 multi-hop      & 16.2\% & 16.0\% & 16.0\% \\
Cat3 temporal       & 4.8\%  & 5.0\%  & 5.0\% \\
Cat4 open-ended     & 42.3\% & 43.2\% & 43.1\% \\
Cat5 counterfactual & 22.5\% & 21.9\% & 21.9\% \\
\bottomrule
\end{tabular}

\vspace{0.75em}

\begin{tabular}{lrrrrrrr}
\toprule
\multicolumn{8}{c}{\textbf{(c) Text and topic diversity indicators}} \\
\midrule
Dataset & Turn Len. & Word TTR & Session Sim. & Adjacent Sim. & 3-gram Repeat & QA TTR & Ref. Len. \\
\midrule
LoCoMo & 124 & 0.0386 & 0.3460 & 0.4537 & 0.316 & 0.1004 & 23 \\
3x     & 137 & 0.0304 & 0.2931 & 0.4253 & 0.319 & 0.0558 & 83 \\
10x    & 136 & 0.0128 & 0.3300 & 0.4496 & 0.451 & 0.0248 & 82 \\
\bottomrule
\end{tabular}

\caption{
Quality analysis of the scaled LoCoMo-style datasets.
Panel (a) reports basic dataset statistics, panel (b) shows that the QA category distribution is preserved across the original and scaled datasets, and panel (c) reports text-level and topic-level diversity indicators.
Turn length and reference length are measured in characters.
Word TTR and QA TTR denote type-token ratios for dialogue turns and questions, respectively.
}
\label{tab:scaled_dataset_quality}
\end{table*}